\documentclass{ieeeaccess}
\usepackage{cite}
\usepackage{amsmath,amssymb,amsfonts}
\usepackage{algorithmic}
\usepackage{textcomp}
\usepackage[pdftex]{graphicx}
\usepackage{enumerate}
\usepackage[shortlabels]{enumitem}
\usepackage{caption}
\usepackage{soul}
\usepackage{hyperref}

\def\BibTeX{{\rm B\kern-.05em{\sc i\kern-.025em b}\kern-.08em
    T\kern-.1667em\lower.7ex\hbox{E}\kern-.125emX}}

\def\eg{\emph{e.g. }}

\def\etal{\emph{et al. }}
\def\ie{\emph{i.e. }}

\begin{document}
	\history{Date of publication December 16, 2019, date of current version January 08, 2020.}
	\doi{10.1109/ACCESS.2019.2961789}

	\title{Rethinking Online Action Detection in Untrimmed Videos: A Novel Online Evaluation Protocol}

	\author{\uppercase{Marcos Baptista-R\'ios}\authorrefmark{1},
			\uppercase{Roberto J. L\'opez-Sastre}\authorrefmark{1},
			\uppercase{Fabian Caba Heilbron}\authorrefmark{2}, 
			\uppercase{Jan C. van Gemert}\authorrefmark{3},
			\uppercase{F. Javier Acevedo-Rodr\'iguez}\authorrefmark{1} and
			\uppercase{Saturnino Maldonado-Basc\'on}\authorrefmark{1}}
		
	\address[1]{GRAM, Department of Signal Theory and Communications, University of Alcal\'a, Alcal\'a de Henares, Spain (e-mail: marcos.baptista@uah.es, robertoj.lopez@uah.es, javier.acevedo@uah.es, saturnino.maldonado@uah.es)}
	
	\address[2]{Adobe Research, Media Intelligence Lab, Deep Learning Group (e-mail: caba@adobe.com)}
	
	\address[3]{Faculty of Electrical Engineering, Mathematics and Computer Science, Delft University of Technology, Delft, The Netherlands (e-mail: j.c.vangemert@tudelft.nl)}
	
	\tfootnote{This work is supported by project PREPEATE, with reference TEC2016-80326-R, of the Spanish Ministry of Economy, Industry and Competitiveness. We gratefully acknowledge the support of NVIDIA Corporation with the donation of a GPU used for this research.}

	\markboth
	{Baptista-R\'ios \headeretal: Rethinking Online Action Detection}
	{Baptista-R\'ios \headeretal: Rethinking Online Action Detection}

	\corresp{Corresponding author: Marcos Baptista-R\'ios (e-mail: marcos.baptista@uah.es).}

	\begin{abstract}
		The Online Action Detection (OAD) problem needs to be revisited. Unlike traditional offline action detection approaches, where the evaluation metrics are clear and well established, in the OAD setting we find very few works and no consensus on the evaluation protocols to be used. In this work we propose to rethink the OAD scenario, clearly defining the problem itself and the main characteristics that the models which are considered online must comply with. We also introduce a novel metric: the Instantaneous Accuracy ($IA$). This new metric exhibits an \emph{online} nature and solves most of the limitations of the previous metrics. We conduct a thorough experimental evaluation on 3 challenging datasets, where the performance of various baseline methods is compared to that of the state-of-the-art. Our results confirm the problems of the previous evaluation protocols, and suggest that an IA-based protocol is more adequate to the online scenario. The baselines models and a development kit with the novel evaluation protocol are publicly available \href{https://github.com/gramuah/ia}{here} (https://github.com/gramuah/ia).
	\end{abstract}
	

	\titlepgskip=-15pt

	\maketitle
	
	\section{Introduction}
	\label{sec:introduction}
	
		\PARstart{I}{n} this work, we focus on the problem of localizing actions in untrimmed videos \emph{as soon as} they happen, which was coined as Online Action Detection (OAD) by De Geest \etal \cite{DeGeest2016}.

		Action detection in video has been widely studied, but \emph{mainly from an offline perspective}, \eg \cite{Shou2016, Shou2017, Gao2017b,Yeung2015, Buch2017b, Dai2017, Buch2017, Chao2018,Xu2017, Zhao2016, Wu2018}, where it is assumed that all the video is available to make predictions. Few works address the \emph{online} setting, \eg \cite{DeGeest2016,Gao2017b,Gao2017c,DeGeest2018}. Think of a robotic platform that must interact with humans in a realistic scenario, or an intelligent video surveillance application designed to raise an alarm when an action is detected. \emph{All} previous offline methods make the described applications impossible because they would detect action situations way later they have occurred.

		\begin{figure*}[t!]
			\centering
			\includegraphics[width=\textwidth]{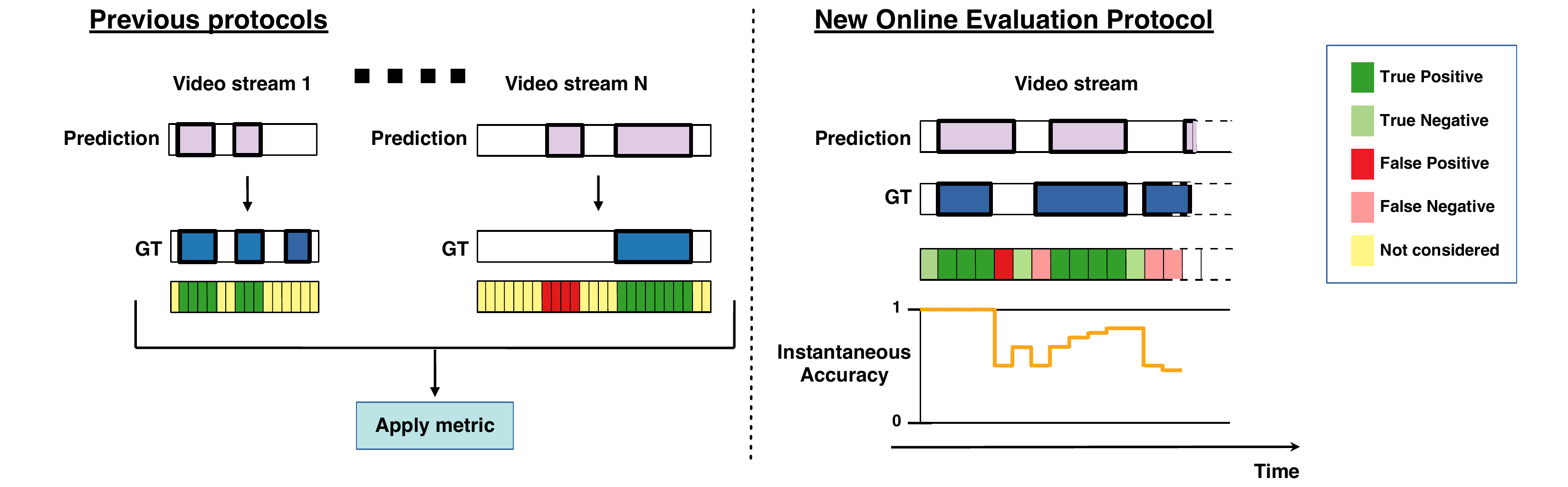}
			\caption{\textbf{Online Evaluation Protocol}. Previous evaluation protocols for Online Action Detection (OAD) were based on: 1) running the online methods through all videos; 2) applying the offline metric on the obtained results. Additionally, offline metrics proposed so far do not consider the background in their evaluation. We propose an Online Evaluation Protocol based on our new Instantaneous Accuracy metric (IA). OAD approaches are evaluated online considering the background and regardless of the length of the video.}
			\label{fig:graphical_abstract}
		\end{figure*}

		On the contrary, in an OAD approach, action detections must be given over video streams, hence working with partial observations, where the action segments are possibly the exception rather than the rule, compared with the background. Moreover, this online definition allows for an important property: the anticipation to the action. In other words, for an OAD model the objective is to anticipate the action even before the action is fully completed.

		However, there are important weaknesses among the online approaches in two fundamental aspects: 1) the evaluation metric; and 2) the treatment of the background category by the models and in the evaluation. Regarding the former, we have noticed that there is no consensus on the evaluation protocols. In each dataset a different metric is proposed for the very same problem. Moreover, used metrics cannot be said to be of an \emph{online nature}. In other words, metrics such as the mean Average Precision (mAP) \cite{Gao2018} or the Calibrated Average Precision (cAP) \cite{DeGeest2016}, do not provide information about the instantaneous performance of the solutions over time. They need to be computed entirely offline, accessing the whole set of action annotations in a given test video, to sort, for instance, all frame predictions. Even the novel point-level Action Start detection mAP metric, proposed in \cite{Shou2018} to evaluate the different problem of online detection of action start, has the same limitation.
		
		Regarding the second aspect, the OAD setting is characterized by long untrimmed videos where actions appear sparsely and the background predominates. Consequently, the online problem should demand the background category to be treated as a first-class citizen. However, if we analyze the online methods published to date, almost all have been designed to cast a specific prediction for the background category: given a test video, every frame is categorized with an action class. For this reason, some propose to modify the evaluation metric, as in ~\cite{DeGeest2016}, where a calibrated version of the average precision is proposed to mitigate the penalty with the background frames. Furthermore, when the background class is not considered in the evaluation, but it is considered in the annotation, all the proposed metrics cannot saturate to the maximum which they have been designed for. In other words, the maximum of a precision-based metric will never be of 100\% even if the method cast for every action frame the correct category.
		
		In this paper we address all the described limitations. Our scientific contributions are as follows:
		\begin{itemize}
			\item First, we introduce an evaluation protocol, with a novel \emph{online} metric: the Instantaneous Accuracy ($IA$)(see Figure \ref{fig:graphical_abstract}). This metric has been designed not only to overcome the limitations, but to allow for fair and effective comparisons between OAD methods.
			\item Second, we propose a thorough experimental evaluation on three challenging datasets (Thumos'14 \cite{THUMOS14}, TVSeries \cite{DeGeest2016} and ActivityNet \cite{Caba-Heilbron2015}), where a comparison between baselines and the state-of-the-art approaches is offered. The results show that an $IA$-based evaluation protocol is more adequate to the OAD problem, because it is able to offer a detailed evolution of the performance of OAD models when the video stream grows over time.
			\item We will publicly release the implementations of the baseline models as well as a development kit with the novel evaluation protocol.
		\end{itemize}
	
	\section{Related Work}
	\label{sec:related_work}

		We summarize here some contributions on related problems: offline action detection, early action detection and online action detection.
		
		\noindent\textbf{Offline Action Detection.} In offline action detection, the whole video is known beforehand and the goal is to detect when and where actions occur. There are works that apply classification on action proposals segments, \eg \cite{Shou2016, Shou2017, Gao2017b, Buch2017, Lin2018, Gao2017}. However, other works \cite{ Yeung2015, Buch2017b, Dai2017,Li2018} train models to directly detect action segments, without the proposal stage. All the previously mentioned works propose fully supervised approaches. Since it is more complicated each day to have labels for such big amount of videos, the community is exploring also weakly supervised alternatives \cite{Zhang2019, Narayan2019, Nguyen2019, Wang2019, Nguyen2018, Shou2018autoloc, Paul2018, zhang2019adapnet}. In any case, our analysis focuses on the different problem of \emph{online} action detection.
		
		\noindent\textbf{Early Action Detection.} In this setting, the objective of the approaches (\eg \cite{Hoai2014,Kong2017,Wang2019}) is to predict the action label of an action video before the ongoing action execution ends. They assume the video stream contains only one action instance and once the video has ended, they decide start and end frames. An Online Action Detection (OAD) scenario makes no assumption on the video and actions must be detected as soon as they happen. F1-score is used for evaluation, but this metric does not meet our online evaluation protocol conditions since it is a class-level metric and background is not considered. 
		
		\noindent\textbf{Online Action Detection.} There exist few recent works on OAD \cite{DeGeest2016, Gao2017c, DeGeest2018, li2016}. De Geest \etal \cite{DeGeest2016} set the OAD conditions and introduced some frame-level baselines models which do not explicitly discriminate action from background. They also proposed two evaluation metrics: a per-frame mean Average Precision and a calibrated Average Precision. In their follow-up work \cite{DeGeest2018}, they designed a two-stream LSTM network to capture better temporal dependencies. Li \etal \cite{li2016} trained a method which predicts skeleton-based action classes (plus background) and regresses the start and end frames. This scenario is simpler and in the lack of an established evaluation protocol, they evaluate their method by adapting traditional offline action detection metrics. Gao \etal \cite{Gao2017c} present a LSTM-based Reinforced Encoder Decoder network which anticipates future frame labels and representations. As a side experiment, they address the OAD task as a special case of anticipation where the anticipation time is set to zero. Therefore this type of network cannot be considered as a pure OAD approach. In our work, we explain why the metrics proposed so far are not suitable for online evaluation in streaming videos and propose a new protocol. We also implement a simple method for OAD capable of explicitly distinguishing action and background. There is also the work of Shou \etal \cite{Shou2018} which focuses on the problem of Online Detection of Action Start (ODAS). ODAS can be seen as a variant of OAD where only the starting point of actions is of interest. An OAD method must always find the start and end of an action. The OAD and ODAS evaluation protocols have in common that both use class-level metrics that are computed offline.

	\section{Online Evaluation Protocol for Online Action Detection}
	\label{sec:oepoad}

		Despite the many practical applications Online Action Detection (OAD) offers, it has been barely explored. As the pioneer work of De Geest \etal \cite{DeGeest2016} stated, OAD needs a solid definition and a strong evaluation protocol, which we revisit in this section. 

		\subsection{Online Action Detection}
		\label{subse:oepoad-oad}

			The established properties of the OAD task in realistic scenarios are summarized as follows: 
			\begin{enumerate}
				\item \textbf{Streaming videos} are assumed, where neither length nor content are known.
				\item Actions must be \textbf{detected as soon as they happen}, ideally in real-time.
				\item \textbf{Detections must be causal.} Future cannot be used, simply because it is not known.
			\end{enumerate}
	
			Note that even though OAD is naturally characterized by untrimmed streaming videos where actions appear sparsely, we found state-of-the-art models that do not consider the background as a category. They treat the OAD problem as a per-frame labeling task where detecting ground truth action frames is what only matters. Mislabeled background frames are dismissed. This means that these methods will not achieve the maximum of a precision-based metric even if the method cast for every action frame the correct category, as we show later in Section \ref{sec:experiments}.
			
			In our exercise of revisting the OAD problem, we propose to add the following properties for OAD methods:
			\begin{itemize}
				\item Methods will explicitly discriminate action from background.
				\item No post-processing or posterior thresholding to action label scores can be applied.
				\item Methods cannot revisit past detections.
			\end{itemize}

		\subsection{Online Evaluation Protocol}
		\label{subsec:oepoad-oep}
		
			A true online evaluation protocol is needed. It is necessary to revisit the evaluation protocol and establish a new one that is in line with the online nature of the OAD problem. We argue an evaluation protocol for OAD must comply with the following conditions: 

			\begin{enumerate}[start=1,label={(\bfseries C\arabic*):}]
				\item An online video-level metric is needed. So method's performance can be evaluated as a video grows without having to wait to an unknown end.
				\item If the OAD task requires methods that are able to detect background, the evaluation protocol must measure such ability.
				\item The value of a \textit{true}, true positive (action) and true negative (background), should be conditioned to the negatives vs. positives ratio, which must be dynamic and based only on the seen portion of the video.
			\end{enumerate}

			\noindent\textbf{Previous metrics.} All previous evaluation protocols use class-level metrics which have to be applied offline, \ie at the end of the test time, accessing the whole set of action annotations in a given test video. Hence, condition (C1) is directly violated. These protocols are mainly based on using the per-frame mean average precision (mAP) or its calibrated version (cAP).

			Regarding mAP, it measures the precision, defined by $Prec = \frac{TP}{TP + FP}$, across all classes. As can be seen in its definition, only positives factors (actions) are considered and their value is always the same regardless of any ratio. Conditions (C2) and (C3) are not complied.

			Precision in cAP is defined by $cPrec = \frac{wTP}{wTP + FP}$. This metric was introduced in \cite{DeGeest2016} and balances the precision with the \textit{w} parameter, which is the ratio between negative vs. positive frames. It is basically a modification of mAP metric so conditions (C1) and (C2) are still not complied. It would solve condition (C3) but \textit{w} is computed a priori (not dynamically) using previous information about all videos and action categories.

			\noindent\textbf{Instantaneous Accuracy metric.} We introduce a new metric which meets all the aforementioned conditions: the Instantaneous Accuracy (IA$(t)$). Considering a set of $\mathcal{N}$ test streaming videos, for each video $\mathcal{V}_i$, where $i=1 \ldots N$, an OAD method generates a set of action detections defined by their initial and ending times. IA metric takes as input these detections to build a dense temporal prediction of background or action for every time slot $\Delta t$ in the test video. Note $\Delta $ is the unique parameter of our IA metric and it measures how often the metric is computed. In section \ref{sec:experiments} we give details on choosing this value.

			For a particular instant of time $0 <t'\leq T_i$, the $\mathrm{IA}(t')$ is computed as the time slot-level accuracy for the classification between action and background:
			\begin{equation}
				\mathrm{IA}(t') = \frac{\sum_{j=0:\Delta t:t'}\vec{tp}(j) + \sum_{j=0:\Delta t:t'}\vec{tn}(j)}{K'} \, ,
				\label{eq:ia}
			\end{equation}
			where $\vec{tp}$ and $\vec{tn}$ are two vectors encoding the true positives (action) and true negatives (background), respectively, according to the predictions and ground truth. $K'$ represents the total population considered until time $t'$, which is dynamically obtained as follows: 
			\begin{equation}
				K' = \left\lfloor \left( \frac{t'}{\Delta t} \right) \right\rfloor\, .
			\end{equation}
			
			To meet condition (C3), and to enable easy and fair comparisons across different OAD approaches, we propose a weighted version of the IA: the wIA. Technically, we scale the \emph{true} factors by the background vs. action slots as follows:
			\begin{equation}
				\mathrm{wIA}(t') = \frac{\sum_{j=0:\Delta t:t'} w(t') \cdot \vec{tp}(j) + \sum_{j=0:\Delta t:t'} \frac{1}{w(t')} \cdot \vec{tn}(j)}{K'} \,
				\label{eq:wia}
			\end{equation}
			where $w(t')$ represents the dynamic ratio between background and action slots until time $t'$ in the ground truth, \ie in $\mathcal{V}_{i}(0:t')$.

			The metric described so far only uses information from the past and is capable of adapting its parameters in each iteration. It would be sufficient to evaluate an OAD method on a single video stream of any length. Additionally, we introduce the mean average Instantaneous Accuracy ($\mathrm{maIA}$) shown in equation \ref{eq:maIA} to summarize a method's performance across a large dataset. In this way, researchers can compare their methods.
			\begin{equation}
				\label{eq:maIA}
				\mathrm{maIA} = \frac{1}{N} \sum_{i=1:N} \left( \frac{\Delta t}{T_i} \sum_{j=0:\Delta t:T_i} \mathrm{IA}(j) \right) \, . 
			\end{equation}
	
	\section{Experiments}
	\label{sec:experiments}

		\subsection{Experimental setup}
		\label{subsec:experiments-setup}

			\begin{figure*}[t]
				\centering
				\includegraphics[width=0.72\textwidth]{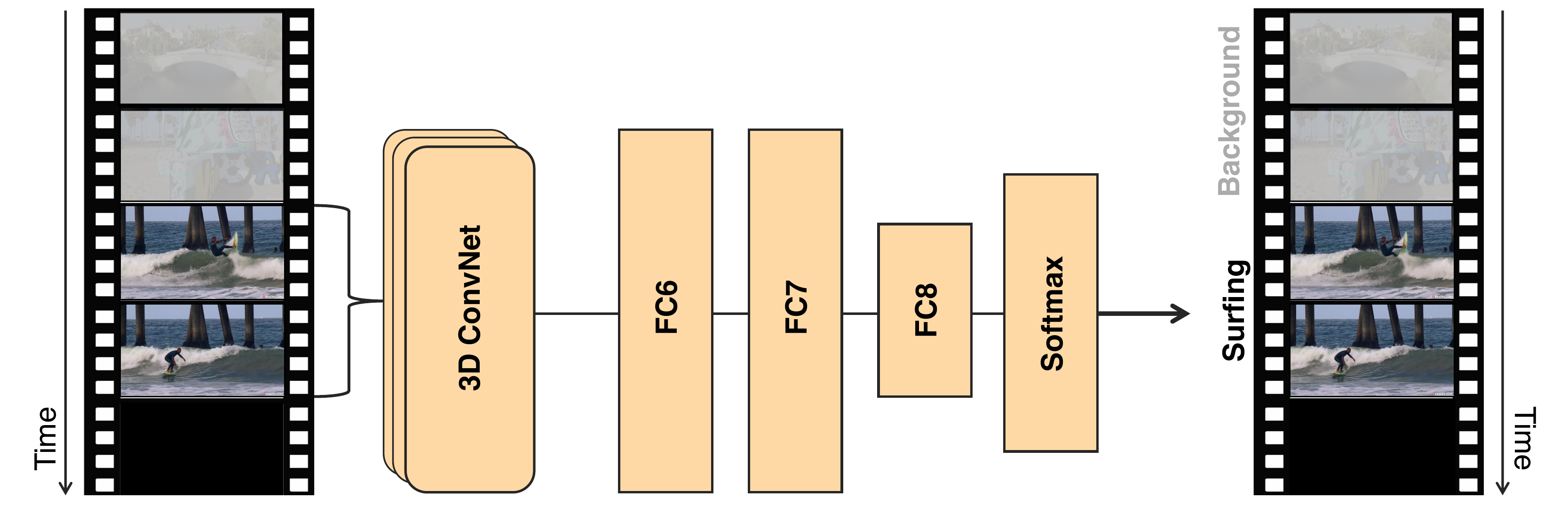}
				\caption{\textbf{3D-CNN baseline model.} Our model closely follows C3D \cite{Tran2015} but trained to discriminate between all action categories plus background. We simply adapted the dimension of the last fully connected layer so that it coincides with the number of categories of interest and the background class.
					The model makes predictions in an online fashion, avoiding to peek into the future for adjusting or post-processing these detections. In short, 3D-CNN generates action and background predictions as the video evolves.}
				\label{fig:esm}
			\end{figure*}

			\textbf{Datasets.} We use three datasets for all our experiments. All of them provide untrimmed videos where action and background segments coexist, suiting our OAD scenario. Thumos'14~\cite{THUMOS14} dataset has temporal annotations for a set of 413 videos, covering 20 sport classes. On average, every video contains 15 action annotations. For training, we use the 200 videos from the validation set, while the remaining 213 from the test set are used for testing. TVSeries~\cite{DeGeest2016} is an OAD-specific dataset. It contains 27 episodes from 6 popular TV series with 30 realistic action categories annotated. Its large variability (occluded, multiple persons or non-relevant actions, among others), makes it a really challenging dataset. Finally, we also integrate in the OAD experiments, ActivityNet v1.3 ~\cite{Caba-Heilbron2015}, which is a large scale dataset specifically designed for Temporal Action Localization. It contains about 20K untrimmed videos for 200 action classes. The average number of action instances per video is of 1.5. For this dataset, we follow the standard procedure: we use the training set and the validation set during training and test respectively. While both Thumos'14 and TVSeries have been already used within the OAD context, we are the first in integrating the challenging ActivityNet into the online setting.
			
			\noindent\textbf{Evaluation Metrics.} On all datasets, we report our novel $IA$ for each video and provide $maIA$ to evaluate methods across each dataset. Following the setup detailed in \cite{Gao2017c}, we analyze the per-frame mAP on Thumos'14 and TVSeries datasets. And finally, for the TVSeries, we analyze the proposed Calibrated Average Precision (cAP) \cite{DeGeest2016}.

		\subsection{Baselines}
		\label{subsec:experiments-baselines}

			In our study, we use three baselines:
			\begin{enumerate}
				\item All background \textbf{(All-BG)}. It simply simulates a model which never outputs an action class, helping to understand the difficulty of the datasets.
				\item Perfect Model (\textbf{PM}), that always assigns correct labels to ground truth action frames and produces a random action label for every background frame. PM helps to reveal the limitations of the mAP and cAP evaluation metrics for OAD, showing they cannot saturate to the maximum.
				\item \textbf{3D-CNN}. As shown in Figure \ref{fig:esm}, it consists of a 3D CNN network trained to discriminate between all action labels plus the background category. Our goal is to establish baseline results for the new online evaluation protocol for OAD with a model capable of explicitly detecting actions and background for the first time.
			\end{enumerate}
			
			Our \textbf{3D-CNN} is based on the C3D network \cite{Tran2015}. Technically, we adapted the dimension of the last fully connected layer of C3D model so that it coincides with the number of classes of interest plus the background category. The architecture is fed with 16-frame length chunks. For training, we extract 16-frame length contiguous chunks. Those whose intersection with ground truth is greater than 0.8 are marked as positive, otherwise they are considered negative (background). The training data $\mathcal{T}$ is balanced by matching the number of samples in each class: $N_{\mathcal{T}} = \frac{N_{chunks}}{\mathcal{C}}$, being $\mathcal{C}$ the total number of classes including background. We initialize our network with Sports-1M \cite{Karpathy2014} weights and SGD is configured with learning rates $10^{-3}$, $10^{-4}$ and $10^{-5}$ for Thumos'14, ActivityNet and TVSeries respectively. For all datasets, momentum is 0.9 and learning rate decreases every 2 epochs. The model is trained for 15 epochs. During test, we simulate the online process on each video by gathering 16 non-overlapping frames and input them to the network, which will cast a prediction. We take the softmax value corresponding to the background class and if it is above 0.8 we consider the detection as background. Otherwise, the detection will be the action class with highest softmax score.
			
			\textbf{3D-CNN} baseline not only is simple but it also requires neither refinement nor post-processing, and can run in real-time (at more than 100 fps). The experimental evaluation shows that it is a strong baseline. Caffe \cite{jia2014caffe} is used for its implementation and it will be publicly available.

		\subsection{A comparison between the metrics}
		\label{subsec:experiments-metrics}

			It is important for us that the reader understands the main weaknesses of the previous evaluation protocols. For this reason, and though it might be unfair, we make a comparison in this section of the performance of all methods with all the metrics on TVSeries dataset.
			
			Table \ref{tab:tvseries_all_metrics} shows the results for all the baselines and the state-of-the-art model in \cite{DeGeest2016}. First, the fact that the PM baseline performance is not the maximum for both cAP and mAP, confirms that using methods and metrics that are not capable of managing the background category is not appropriate for the OAD task. Even though the cAP metric seems to alleviate this problem, it is not enough to achieve a 100\% and it is based on diminishing background errors.
			Second, All-BG baseline reveals: a) that previous metrics are not able to measure method's ability to distinguish both action and background and b) the need of having a metric such as IA, capable of weighting the relevance of errors in both action and background. This last fact is especially important when dealing with very unbalanced datasets like TVSeries.
			Third, results from 3D-CNN are competitive when compared to the state-of-the-art. So it is confirmed as a strong baseline for the OAD problem. It is only with the cAP metric that CNN \cite{DeGeest2016} really outperforms it. The reason is that this method does not cast predictions of background category (while 3D-CNN does) and, as said before, cAP has been designed to minimize the importance of such errors.
			
			\begin{table}[t!]
				\centering
				\caption{Analysis of all the metrics on TVSeries.}
				\resizebox{7cm}{!}{
					\begin{tabular}{c|c|ccc}
						& CNN \cite{DeGeest2016} & All-BG  & 3D-CNN & PM  \\ \hline
						mAP (\%) & 1.9 & 0 & 1.6 & 30.9 \\ 
						cAP (\%) & 60.8 & 0 & 10.8 & 96.9 \\ \hline
						maIA (\%) & 3.51 & 78.3 & 71.9 & 100 \\
						weighted maIA (\%) & 12.46 & 22.9 & 28.9 & 100 \\ \hline
					\end{tabular}
				}
				\label{tab:tvseries_all_metrics}
			\end{table}
			
			In the OAD problem, it is fundamental to consider the background as one more category in the video. While our 3D-CNN baseline does explicitly consider it, most state-of-the-art online models do not. How does this fact affect performances? We analyze this on Thumos'14. 
			
			Table \ref{tab:map-th14}, shows the per-frame mAP achieved by all state-of-the-art models and our 3D-CNN. 
			The poor performance of the perfect model confirms again the limitation of the mAP metric. Additionally, 3D-CNN results on this dataset also demonstrate this model is a good baseline for OAD.
			It is important noticing that all state-of-the-art methods assign an action category to every frame in the video, including those frames that belong to background segments. Moreover, the metric is not considering background errors. This means that mAP does not encourage methods to correctly discriminate background segments.
			To be precise, RED \cite{Gao2017c} does use the background during training to predict sequences of labels. But at test time, in no case is the background separated from the action. Furthermore, RED is designed for anticipation and these results are obtained when taking a very short anticipation time. So, it cannot be considered as a pure online action detection method, because it violates the causality condition. In any case, from this perspective, the performance reported by 3D-CNN is even more relevant: while our model has been trained to deal with a harder problem, it is able to maintain a state-of-the-art performance.
			
			\begin{table}[t!]
				\centering
				\caption{Per-frame mAP performance on Thumos'14.}
				\resizebox{7cm}{!}{
					\begin{tabular}{c|ccc|cc}
						& TS-CNN\cite{Yeung2018} & MultiLSTM \cite{Yeung2018} & RED \cite{Gao2017c}  & 3D-CNN  & PM  \\ \hline
						mAP (\%) & 36.2   & 41.3      & 45.3 & 30.1 & 57.0
					\end{tabular}
				}
				\label{tab:map-th14}
			\end{table}
			
			Finally, we want to emphasize that neither mAP nor cAP are \emph{online} metrics. Results in Tables \ref{tab:map-th14} and \ref{tab:tvseries_all_metrics} for these two metrics can only be reported once the methods have been executed on all the videos.
			Instead, our IA metric is online. It can perform a true online comparison between OAD models, as we show in the next section.
			
			Overall, we conclude that a novel online metric with an adequate evaluation protocol is needed.

		\subsection{Evaluation with Instantaneous Accuracy}
		\label{subse:experiments-ia}

			We analyze our IA metric with the 3D-CNN baseline and the CNN \cite{DeGeest2016} approach. We have not found any code or results of other OAD state-of-the-art methods, except for CNN \cite{DeGeest2016} and LSTM \cite{DeGeest2016}. Since the performance of CNN \cite{DeGeest2016} and LSTM \cite{DeGeest2016} are similar, we have decided to generate the results of the first for its simplicity. We have exactly reproduced the code provided by the authors. Note that these methods do not recognize background. Additionally, we tried to use results from offline temporal action detection methods but did not find a fair adaptation of them.
			
			\noindent\textbf{Instantaneous Accuracy for evaluation in online streaming videos.} In a nutshell, our novel IA measures in an online way how accurate a certain OAD method is being along the streaming video, based only on what has been seen up to the instant of evaluation.
			
\begin{figure}[t!]
				\centering
				\includegraphics[width=\columnwidth]{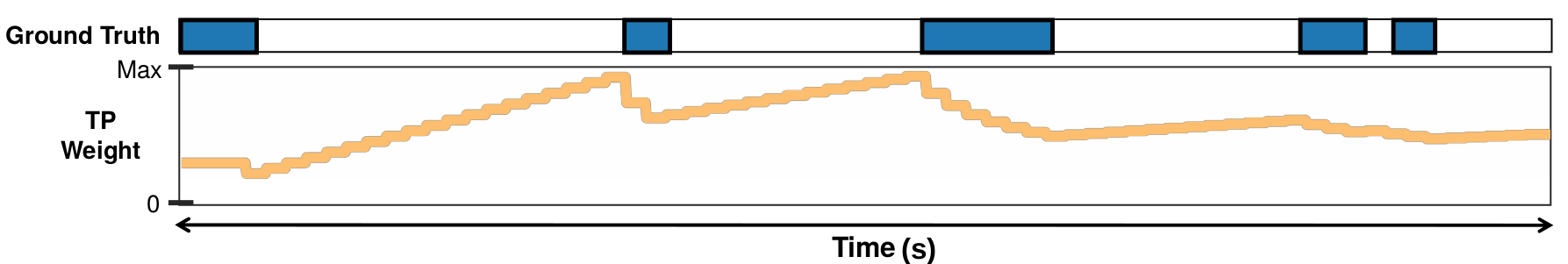}
				\caption{\textbf{Dynamic weights.} It can be seen how the value of a True Positive (TP) is weighted according to the ratio of negative/positive slots seen up to the instant of evaluation. As Equation \ref{eq:wia} shows, TN weights offer the inverse effect.}
				\label{fig:weights}
			\end{figure}

			Regarding the parameters of the metric, the slot duration represents how often the evaluation is applied and it should be 0. Since such an ideal duration is technically not achievable, we have configured it to be the shortest possible. Most action detection approaches use chunk-level features. The chunk length is typically in the interval $[16,64]$ frames,  with a 25/30 frame rate, representing each chunk about 0.6 to 2 seconds. Thus, we choose a 0.5 seconds for slot duration parameter $\Delta t$. The IA metric considers correct predictions of both true positives (action categories) and true negatives (background). The value of a true prediction is dynamically weighted according to the ratio of negative/positive slots seen so far. Figure \ref{fig:weights} shows this dynamic behaviour of the weights. The weight applied to TP (the weight of the TN is the inverse) changes throughout the video. Action and background are not always balanced at each instant of evaluation during the video stream. For this reason, the weight of the true positive predictions (finding action) increases in those portions of the video in which there is no action annotated. This fact represents how the IA metric is modulating the importance of a good prediction and it is a very relevant difference with previous protocols.
			
\begin{figure}[h!]
				\centering
				\includegraphics[width=1\columnwidth]{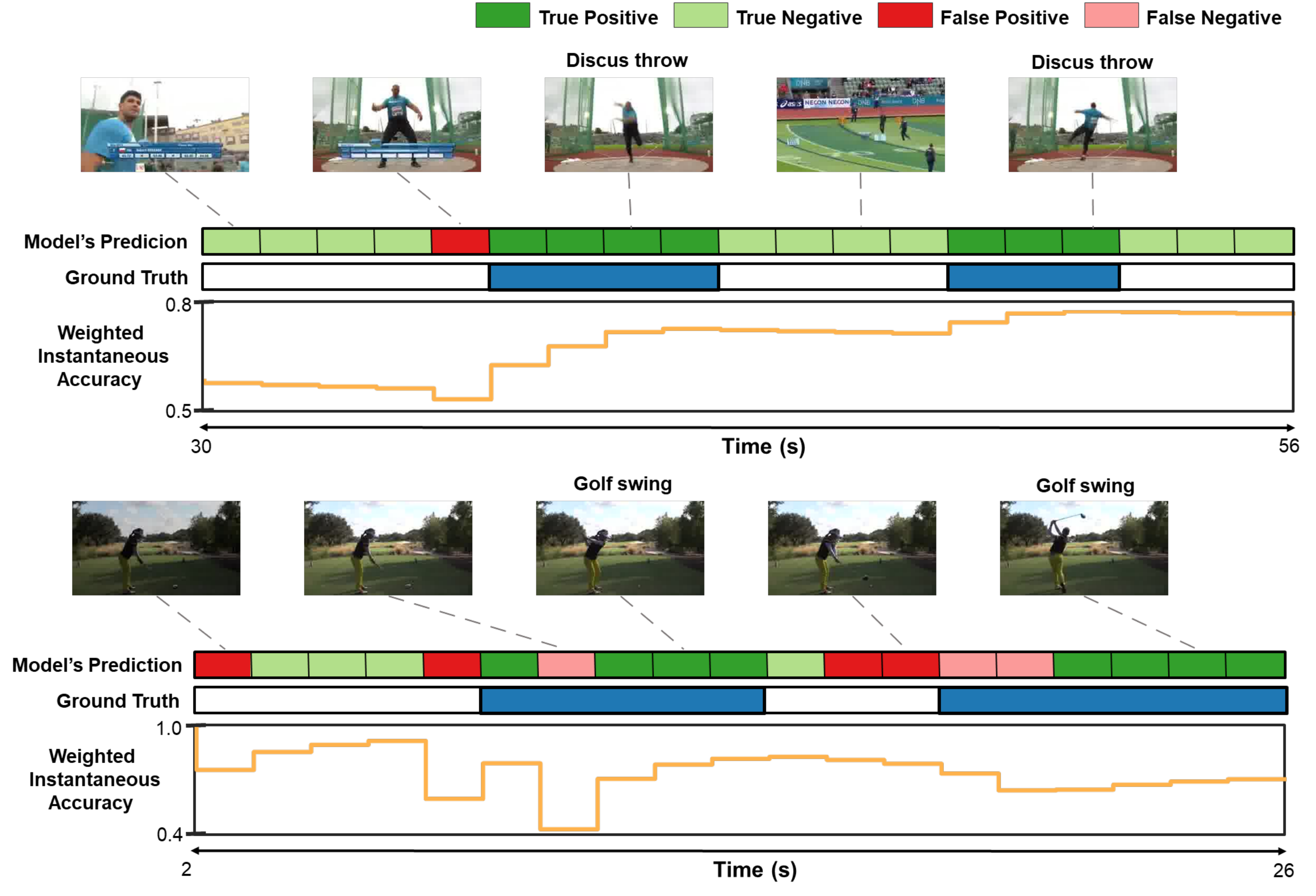}
				\caption{\textbf{Qualitative Results.} We showcase the evolution of the weighted IA on two different Thumos'14 videos. Each instant of evaluation depends on the current model's prediction. IA metric is an online video-level metric which measures the ability of methods to discriminate actions and background.}
				\label{fig:qualitative-results}
			\end{figure}

			In Figure \ref{fig:qualitative-results} we show qualitative results for two different videos. Only a section of the videos is shown. Note that an accuracy value for a certain slot does not depend on that of the previous slot. Those values rely only on the predictions and the weights for each correct prediction. These dynamic weights can lead to situations where the accuracy value decreases (not much) even if a method is getting right predictions. This effect is seen in the upper example of Figure \ref{fig:qualitative-results}, in the segment between the ground truth annotations. However, that is exactly what we want: since nothing about the video is known before, the importance of detecting action or background must vary throughout the streaming.
			
\begin{figure*}[t]
				\begin{center}
					\includegraphics[width=1\linewidth]{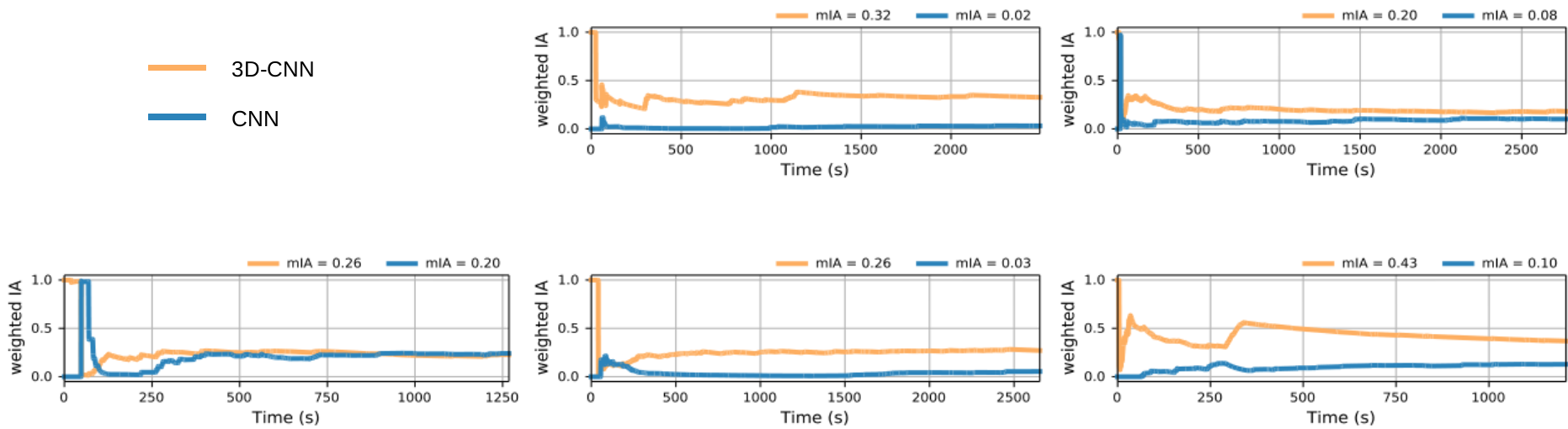}
				\end{center}
				\caption{Online IA based evaluation in videos of the TVSeries dataset.}
				\label{fig:ia_evolution}
			\end{figure*}

			Figure \ref{fig:ia_evolution} shows the evolution of the weighted IA on the 7 videos of the test subset of TVSeries dataset. Note how it allows for a true online comparison between OAD models, in this case, CNN \cite{DeGeest2016} and 3D-CNN.
			
			\noindent\textbf{maIA as IA consolidation for evaluation across datasets.} Despite the fact that IA metric can be directly used as it is in a video stream, we propose also the maIA to compare methods on a certain dataset. 
			
			Table \ref{tab:maia} presents the performance with the weighted and non-weighted versions of maIA on the three datasets. Results from All-BG baseline reveal the relevance of having a weighted metric. Thumos'14 and TVSeries are very unbalanced datasets and when introducing the weighting, the performance drops a lot. On ActivityNet All-BG performs similar with the two versions of the metric due to the fact that the dataset is more balanced. These results confirm the consistency of our metric, which is capable of making a fair evaluation in all kind of datasets.
			
			The low numbers of 3D-CNN on TVSeries and ActivityNet are caused by different reasons. TVSeries is a specially very unbalanced and challenging dataset. With such a lot of background, a model as simple as 3D-CNN is not able to learn well to discriminate action from background. ActivityNet is balanced but has many classes to distinguish. Finally, our reproduced CNN \cite{DeGeest2016} performs poorly according to maIA due to it does not handle background. Thus, its performance is alleviated when weighted with the positive/negative ratio.
			
			\begin{table*}[htpb]
				\centering
				\caption{Weighted an non-weighted maIA on Thumos'14, TVSeries and ActivityNet.}
				\label{tab:maia}
				\begin{tabular}{c|cc|ccc|cc}
					& \multicolumn{2}{c|}{\textbf{Thumos'14}} & \multicolumn{3}{c|}{\textbf{TVSeries}} & \multicolumn{2}{c}{\textbf{ActivityNet}} \\ \cline{2-8} 
					& All-BG              & 3D-CNN               & All-BG        & 3D-CNN        & CNN \cite{DeGeest2016}      & All-BG                 & 3D-CNN                 \\ \hline
					maIA (\%)          & 70.9                & 72.64              & 78.3          & 71.9       & 3.51         & 40.1                   & 21.7                \\
					weighted maIA (\%) & 41.8                & 58.10              & 22.9          & 28.9       & 12.46         & 53.6                   & 27.4               
				\end{tabular}
			\end{table*}

	\section{Conclusion}
	\label{sec:conclusion}

		Online Action Detection in untrimmed streaming videos is a challenging task with few contributions. We have found that a) the task itself needs a solid definition of its properties, b) there is no clear consensus on how the methods should deal with this type of videos and c) a proper evaluation protocol is not defined.
		
		In this work, we solved the first two problems by revising and establishing the properties of the OAD task itself as well as those for the methods designed for it.
		Regarding the third, we noticed that there are limitations in the metrics used so far. Therefore, we have clearly defined the conditions of a proper evaluation protocol: i) it has to be online, for consistency with the metric; ii) it must measure the ability of methods to discriminate both action and background and iii) it must be based only on the seen portion of video.
		
		Since none of the previously used metrics complies with these conditions, we have introduced a new metric: the Instantaneous Accuracy (IA). IA is an online video-level metric which computes the accuracy for every instant of evaluation. Our results have proved the limitations of the previous metrics and the robustness of our novel IA.
		
		We expect in the future more methods will be analyzed with our IA. Thanks to its characteristics, it will be possible to study the situations in which methods should perform better.
		
		The baselines models and a development kit with the novel evaluation protocol are publicly available \href{https://github.com/gramuah/ia}{here} (https://github.com/gramuah/ia). 

	\bibliographystyle{IEEEtran}
	\bibliography{egbib}
	
	\EOD

\end{document}